\documentclass[conference]{IEEEtran} 
\usepackage{newtxtext}
\usepackage{cite}
\usepackage{amsmath,amssymb,amsfonts}
\usepackage{algorithmic}
\usepackage{graphicx}
\usepackage{textcomp}
\usepackage{xcolor}
\def\BibTeX{{\rm B\kern-.05em{\sc i\kern-.025em b}\kern-.08em
		T\kern-.1667em\lower.7ex\hbox{E}\kern-.125emX}}

\usepackage[symbol]{footmisc}
\usepackage[algoruled,vlined,linesnumbered]{algorithm2e}
\usepackage{booktabs}
\usepackage{latexsym} 
\usepackage{units}
\usepackage{MnSymbol}
\usepackage{comment}
\usepackage{extarrows}
\usepackage{diagbox}
\usepackage{qtree}
\usepackage{bm}
\usepackage{fancyvrb} % for "\Verb" macro
\usepackage{makecell} % for multi-lines across tables 
\usepackage{amsthm}
\theoremstyle{definition} 
\newtheorem{thm}{Theorem}
\newtheorem{thm2}{Theorem}
\newtheorem{cor}{Corollary}

\usepackage{fancyvrb} % for "\Verb" macro

\usepackage{color} % to be used only for comments

\newtheorem{df}{Definition}

\newtheorem{theorem}{Theorem}
\newtheorem{lemma}{Lemma}%[section]
\newtheorem{col}{Corollary}

\newcommand{\bt}{\begin{theorem}\em}
	\newcommand{\et}{\end{theorem}}
\newcommand{\bl}{\begin{lemma}\em}
	\newcommand{\el}{\end{lemma}}
\newcommand{\bc}{\begin{col}\em}
	\newcommand{\ec}{\end{col}}

\renewcommand{\nin}{\noindent}

\newcommand{\bea}{\begin{eqnarray}}
	\newcommand{\eea}{\end{eqnarray}}

\newcommand{\bdf}{\begin{df}\em}
	\newcommand{\edf}{\end{df}}

\newcommand{\ben}{\begin{enumerate}}
	\newcommand{\een}{\end{enumerate}}

\newcommand{\argmax}{\operatornamewithlimits{argmax}\limits}

\usepackage{listings}
\title{Solving Witness Triangle Puzzles Faster with an Automatically Learned Predicate}

\author{
    \IEEEauthorblockN{Justin Stevens\IEEEauthorrefmark{1}, Vadim Bulitko\IEEEauthorrefmark{1}, and David Thue\IEEEauthorrefmark{2}}
    \IEEEauthorblockA{\IEEEauthorrefmark{1}Department of Computing Science, University of Alberta}
    \IEEEauthorblockA{\IEEEauthorrefmark{2}School of Information Technology, Carleton University}
}

\newcommand\ouralg{\text{A}^*_\pi}

\begin{document}
	
	\maketitle
	
	\begin{abstract}
		\textit{The Witness} is a game with combinatorial puzzles that are challenging for both human players and AI solvers. 
        Search-based AI solvers can require large amounts of time to solve such puzzles.
        We accelerate search by automatically learning a predicate that predicts whether a partial path to a Witness puzzle is  
        completable to a solution path. 
        We demonstrate that our learned predicate is human-explainable, which makes it amenable to proofs, and we prove a key property that allows us to use it for pruning successor states in search.
        Doing so accelerates search by about six times while maintaining search completeness. % of the underlying search. 
		Conversely, given a fixed time budget per puzzle, our predicate-accelerated search can solve more puzzles of larger sizes than a baseline search using a hand-coded predicate. 
        Finally, we empirically demonstrate how accelerating search using an automatically learned predicate can outperform two competitors, weighted A* and Levin tree search with neural networks, in terms of time and number of nodes expanded. 
	\end{abstract}
	
	\begin{IEEEkeywords}
		heuristic search, program synthesis, machine learning, inductive logic programming 
	\end{IEEEkeywords}
	
    %%%%%%%%%%%%%%%%%%%%%%%%%%%%%%%%%%%%%%%%%%%%%%%%%%%%%%%%%%%%%%%%%%%%%%%%%%%%%%%%%%%%%%
	\section{Introduction}
	
	{\em The Witness} is a commercial video game with constrained combinatorial pathfinding puzzles~\cite{witness}. 
    In the game, triangle puzzles are scattered throughout the open world environment for players to find and solve. 
    They also play a part in the final challenge to players. 
    Solving its puzzles automatically with artificial intelligence (AI) can be useful to both players and puzzle designers, as they can use such a solver to see a solution to a puzzle. 
    Designers can also use AI solvers in concert with a procedural puzzle generator to generate puzzles with desired solution qualities~\cite{de2019procedural}. 
	
	Various search techniques can be applied to Witness puzzles. 
    For instance, one way to solve Witness puzzles (Figure~\ref{fig:expuzzle}) is to use a complete search that enumerates all possible solutions~\cite{NathanWitnessSolver}.
    However, for puzzles of larger sizes such enumeration becomes intractable. 
    For instance, for the puzzle size of $8 \times 8$, the number of candidate solution paths from the bottom left to the top right corner is on the order of $10^{15}$~\cite{iwashita2012zdd}. For the type of Witness puzzles we consider in this paper, an arbitrary-sized version of them is NP-complete~\cite{abel2020witnesses}. 
    
	To scale up search to larger instances of Witness puzzles, we automatically learn a predicate that predicts whether a partial path considered during search can be completed to a solution path. 
    Accurate predicates can help prioritize the exploration of more promising partial paths. 
    We automatically learn candidate predicates from training data using an off-the-shelf Inductive Logic Programming (ILP) system, {\em Popper}~\cite{cropper2021learning}. 
    The learned predicates are human readable, which can make them amenable to proving their correctness.
    We assess how much the learned predicates can speed up the search for puzzle solutions in terms of time and nodes expanded, and select the fastest candidate.
    We compare the effectiveness of our machine-learned predicate with a human-designed predicate that implements a simple strategy for pruning partial paths.
    We also compare an A* search that uses our machine-learned predicate to two other approaches: weighted A*~\cite{pohl1970first} and Levin tree search with a neural network~\cite{orseauL21}.

	We offer a novel way to learn predicates that can accelerate a search for Witness puzzle solutions. 
    Our best learned predicate was human-explainable, so we could prove a key property of its behaviour. This lets us use it for pruning while maintaining the completeness of the search. 
    Using a testing set of $15,000$ puzzles, we show that our predicate-accelerated search can speed up search more than weighted A* and Levin tree search with neural networks. 
	Finally, we make our code and data available to the community upon request, with the goal of introducing Witness puzzles as a standard benchmark for researchers studying artificial intelligence for games.

	%%%%%%%%%%%%%%%%%%%%%%%%%%%%%%%%%%%%%%%%%%%%%%%%%%%%%%%%%%%%%%%%%%%%%%%%%%%%%%%%%%%%%%
	
	\section{Background}
	\label{sec:problemFormulation}
	We first formally define the kind of Witness puzzles that we consider in this paper. 
    We then describe how a heuristic search for puzzle solutions can be accelerated using a predicate. 
    Finally, we describe performance measures to test the efficacy of these predicates.
	
	\subsection{The Witness Triangle Puzzle}\label{sec:definition}
	
    In this section, we formally describe the rules of triangle puzzles from the game. 

       \begin{figure}[htb]
		\centering
		\includegraphics[width=0.8\columnwidth]{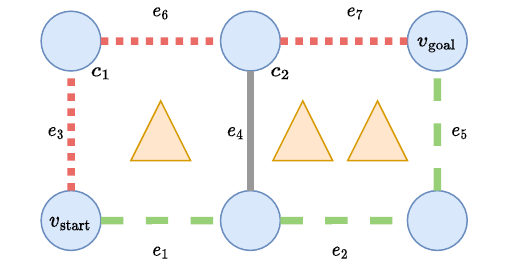}
		\caption{A $1\times 2$ triangle puzzle. The green dashed line solves the puzzle since the solution path intersects the left square once and right square twice, matching the numbers of triangles in their respective squares. The red dotted line is not a solution to the puzzle.}	
		\label{fig:expuzzle}
	\end{figure}
	
	An $m \times n$ {\em Witness-type triangle puzzle} instance $p=(G, v_\text{start}, v_\text{goal}, C)$ is a single-player combinatorial puzzle that uses the constraint in the triangle puzzle from \textit{The Witness}; henceforth we refer to them as \textit{puzzles}. 
	
	Puzzles are solved on a two-dimensional rectangular grid with $m\times n$ squares. 
	Here, $G=(V, E)$ is a graph representing the grid. 
	Each $v \in V$ is a vertex on the grid containing a corner of a puzzle square.
	An $m \times n$ puzzle is thus represented by an $(m+1) \times (n+1)$ rectangular grid of vertices and edges. 
	To illustrate, the $1 \times 2$ puzzle in Figure~\ref{fig:expuzzle} has one row of two squares. 
	It is represented by a $2 \times 3$ rectangular grid with two rows of three vertices each (shown as the blue circles). There are seven edges, $e_1, \dots, e_7$,  connecting the six vertices.

	Each {\em square} is delineated by four grid edges and can carry a {\em constraint} $c \in C$ which associates that square with a number of {\em triangles} contained in that square. 
    A \textit{solution} to the puzzle is a path on $G$ which (i) connects the start vertex $v_\text{start}$ to the goal vertex $v_\text{goal}$ using the edges in $E$, (ii) never visits a vertex more than once, and (iii) satisfies all constraints in $C$. 
    To satisfy a square's constraint $c \in C$ (i.e., the $k > 0$ triangles in that square) the path must include exactly $k$ of that square's four edges. 
    We say that a path \textit{intersects} a square if the path includes at least one of the square's edges. A square without triangles imposes no constraints. 
	
	To illustrate, the $1 \times 2$ puzzle in Figure~\ref{fig:expuzzle} has two constraints. 
    The left square contains a single triangle ($k = 1$) which means that any solution path to the puzzle must include exactly one of the square's edges. 
    The right square carries two triangles ($k = 2$), requiring two of that square's edges in any solution path. The red dotted line connecting the start and goal vertices in Figure~\ref{fig:expuzzle} is not a solution since it violates both constraints (i.e., includes two edges from the first square and only one edge from the second square). 
    The dashed green line is a solution that satisfies both constraints. 
    A {\em partial path} is any path in $G$ that starts out in $v_\text{start}$ but does not reach $v_\text{goal}$. 
    A partial path is {\em incompletable} if it is not a prefix of a solution.
    In Figure~\ref{fig:expuzzle}, the partial path $[e_3, e_6]$ is incompletable while $[e_1, e_2]$ is \textit{completable}, meaning it can be extended to a solution.
    {\em Solution length} is the number of edges in the solution.

	\subsection{Search for Solving Puzzles}
	\label{sec:searchAlgorithm}
	
	We now describe an A* search algorithm~\cite{hart1968formal} applied to solving puzzles (Algorithm~\ref{alg:a*}). 
    Let $\mathcal{P}$ be the space of all puzzles and $\mathcal{L}$ be the space of all partial paths on the puzzles.
    We add a function $\pi: \mathcal{L}\times \mathcal{P}\to \mathbb{R}$ to the algorithm.
	
	The search starts at $v_\text{start}$ and proceeds by expanding its search frontier (i.e., the open list) containing partial paths and housed in a priority queue. 
    The priority queue is sorted by $(\pi, g+h, h)$, where $g$ is the length of the partial path so far, and $h$ is a heuristic estimate of the remaining path length to the solution. 
    First, the queue is sorted such that partial paths with lower values of $\pi$ are returned from the queue first. 
    Next, any ties among partial paths based on their $\pi$ values are broken in favor of lower $g+h$. 
    Finally, any remaining ties are broken in favor of lower $h$, which is equivalent to breaking ties towards higher $g$, a common technique in A* search. 
    Any residual ties are then broken in an arbitrary fixed order. Given the four-connected rectangular grid underlying our puzzles, we use Manhattan distance (MD) as the heuristic $h$. 
	
	\begin{algorithm}[htbp]
		\DontPrintSemicolon
		\caption{A* search for triangle puzzles.}
		\label{alg:a*}
		\SetKwInOut{Input}{input}
		\SetKwInOut{Output}{output\,}
		\Input{puzzle $p=(G, v_\text{start}, v_\text{goal}, C)$, predicate $\pi$}
		\Output{solution path $\ell$}
		
		$Q \gets \text{priority queue sorted by } (\pi, g+h, h)$\; \label{al:sort}
		
		push $v_{\text{start}}$ onto $Q$\; \label{al:init}
		
		\While{$Q \neq \emptyset$}{\label{al:loop}
			$\ell \gets \text{head}(Q)$\;  \label{al:pop}
			$Q\gets Q\setminus \{\ell\}$\; \label{al:pop2}
			$v \gets \text{\rm end}(\ell)$\;  \label{al:head}
			
			\ForEach{$v_{\text{\rm new}} \in N(v) \setminus \ell$}{ \label{al:neighbour}
				$\ell_{\text{\rm new}} \gets [\ell, v_\text{\rm new}]$ \label{al:newlist}\;
				
				\uIf{$v_{\text{\rm new}} = v_\text{\rm goal}$}{ \label{al:goal} 
					\If{$\forall c \in C \left[ \text{\rm $\ell_{\text{\rm new}}$ satisfies $c$} \right]$}{ \label{al:constraint}
						\Return $\ell_{\text{\rm new}}$\; \label{al:solutionFound}
					}
				}
				\Else{	
					push $\ell_{\text{\rm new}}$ onto $Q$\; \label{al:pushNew}
				}
			}
			
		}
		\Return $\emptyset$ \;
	\end{algorithm}
	
	The search loop continues as long as the open list is not empty (line~\ref{al:loop}) and a solution is not found (line~\ref{al:solutionFound}). 
    The first sorted element of the open list, a partial path $\ell$, is retrieved and removed from the queue in lines~\ref{al:pop} and \ref{al:pop2} and the last vertex $v$ of the path is retrieved in line~\ref{al:head}.
    We then expand the end of the path $v$ by generating all neighbours $v_{\text{new}}$ of $v$ in the graph $G$ that are not already on the path $\ell$ (line~\ref{al:neighbour}). 
    We generate each new path by appending $v_{\text{new}}$ to the end of $\ell$ (line~\ref{al:newlist}). 
    If the new path $\ell_{\text{\rm new}}$ reaches the goal vertex, we check if it satisfies all constraints in $C$ (line~\ref{al:constraint}).
    If so, the search stops and the solution is returned in line~\ref{al:solutionFound}. 
    Otherwise, each new partial path is put onto the open list with its $\pi, g+h,$ and $h$ values (line~\ref{al:pushNew}). 
    When $\pi$ is constant we have standard A*.

    We restrict $\pi$ to a predicate $\pi:  \mathcal{L}\times \mathcal{P}\to \{\text{True}, \text{False}\}$ 
    Such a predicate takes a partial path and a puzzle instance, returning True if it predicts that the partial path is incompletable to a solution, and False if it predicts that the partial path is completable. In practice, we compute $\pi$ on each individual constraint, and if it violates a single constraint from the entire puzzle, then we predict the partial path to be incompletable.
    In sorting by $\pi$, partial paths for which $\pi$ returns False (completable) are placed earlier in the queue than paths labeled True (incompletable). 
    An accurate predicate will thus focus the search on completable partial paths. 
    
\begin{figure}[htb]
		\centering
		\includegraphics[width=0.99\columnwidth]{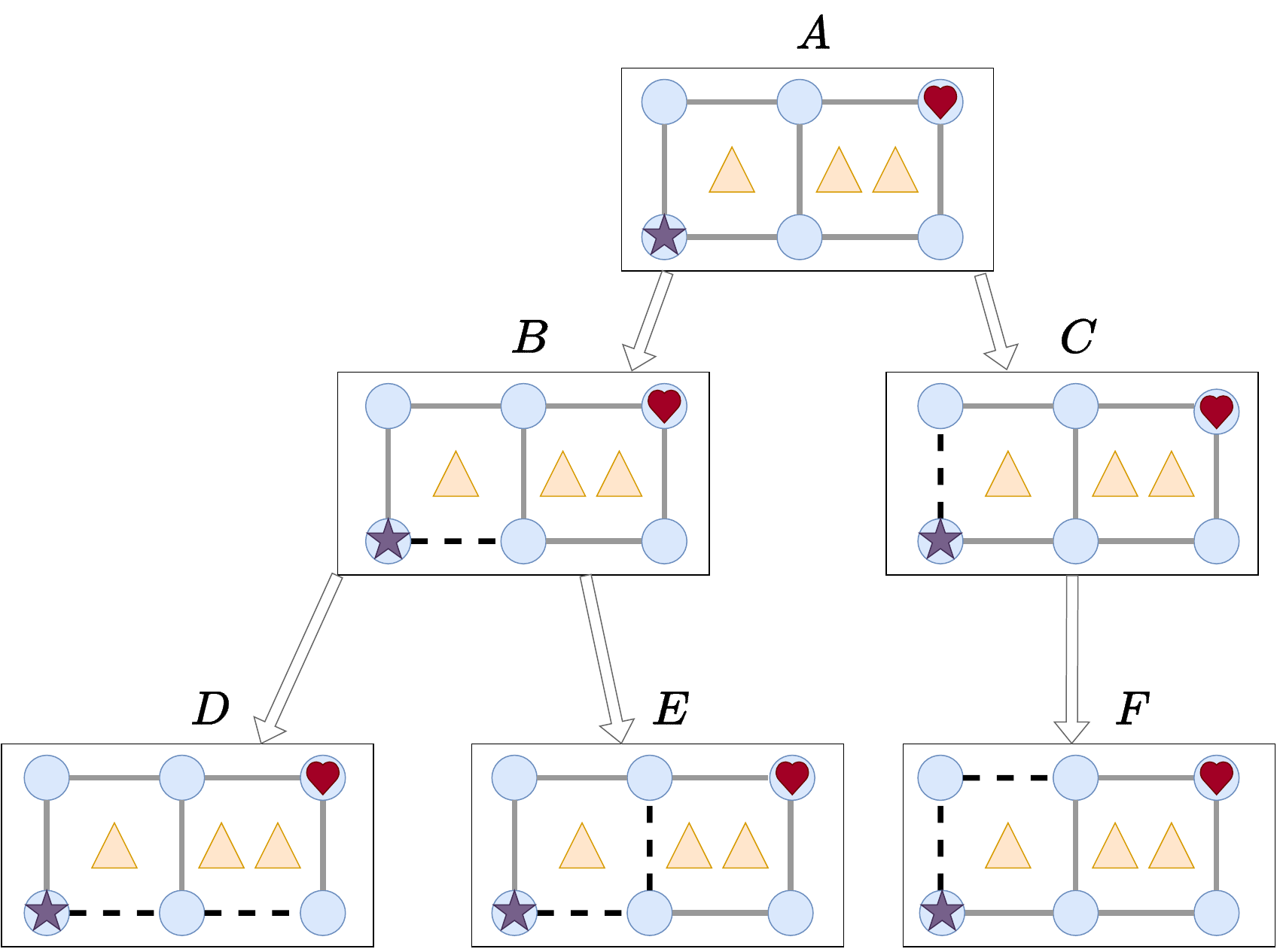}
		\caption{A partial search tree for solving the puzzle in Figure~\ref{fig:expuzzle}. The dashed lines show the partial paths for solving the puzzle so far. The two vertex markers $\star, \heartsuit$ mark the start and goal vertices of the grid.}	
		\label{fig:exsearchtree}
	\end{figure}

    To illustrate, consider the search tree in Figure~\ref{fig:exsearchtree}. Imagine we had a predicate that predicts a partial path is incompletable if it has more intersections than triangles in a square. 
    The A* search with such a predicate would expand node $D$ and its children before expanding nodes $E$ or $F$. 
    
	Note that Algorithm~\ref{alg:a*}'s search is {\em complete} (i.e., it will find a solution to a puzzle if there is one) even when $\pi$ is not completely accurate, since any solution that is wrongly predicted to be incompletable by $\pi$ will still end up in the queue -- just with a lower priority. 

    \subsection{Pruning by $\pi$}

	We can also consider pruning the open list ($Q$) by $\pi$ instead of simply sorting by $\pi$. 
    When using $\pi$ for pruning, partial paths for which $\pi$ returned True (incompletable) are not added to $Q$.
    Although pruning by $\pi$ risks losing completeness of the search, Theorem 1 shows how this risk is avoided when $\pi$ has no false positives (i.e., when it never predicts that a completable partial path is incompletable).  
    See Appendix~\ref{sec:proofs} for all proofs.

       \begin{thm} When $\pi$ has no false positives pruning partial paths that $\pi$ predicts are incompletable preserves completeness of the search. \label{thm:fp} \end{thm}
       
	\subsection{Baseline} \label{sec:intuition}
	
	\nin To establish a basis of comparison for the search that we accelerated using an automatically-learned predicate, we used our knowledge of the puzzle to design a predicate by hand. 
	Specifically, we noted that any partial path that includes more of a square's edges than there are triangles in that square cannot possibly be completed to a full solution. We call the predicate that predicts completability based on this observation \textit{local constraint checking}.

	To illustrate, consider the search tree in Figure~\ref{fig:exsearchtree} for solving the puzzle in Figure~\ref{fig:expuzzle}. 
    The predicate would label partial paths $E$ and $F$ incompletable since they include two edges of the square that has only one triangle in it.

	\begin{thm} The local constraint checking predicate has no false positives.\label{thm:lcc} \end{thm}

	\begin{cor} \sloppy We can use local constraint checking to prune partial paths while preserving search 
    completeness.\label{cor:lcc} \end{cor} 

	Consequently, of the partial paths of length two (the bottom row in the figure), only $D$ would be added to the priority queue since only this partial path is a solution prefix.
    Let $\pi_{\text{baseline}}$ be the local constraint checking predicate. 
    We use Algorithm~\ref{alg:a*} with pruning by $\pi_{\text{baseline}}$ as our \textit{baseline algorithm}.

	\section{Problem Formulation}\label{sec:accel} 
	
	The problem we tackle in this paper is to automatically learn a predicate $\pi$ that speeds up Algorithm~\ref{alg:a*} with pruning on a set of puzzles $\mathcal{P}=\{p_1, \ldots, p_n\}$, relative to our baseline. 
 
    We use two metrics to quantify the speed-up: improvements in the {solution time} and the {number of nodes expanded}. 
	
	For a given puzzle $p\in \mathcal{P}$, let $t(a, \phi, p)$ be the {\em time} it takes an algorithm $a$ with a parameter $\phi$ to solve that puzzle. 
    Similarly, let $\mathcal{E}(a, \phi, p)$ be the {\em number of nodes expanded} by $a$ as it solves $p$ using parameter $\phi$. 
    The set of algorithms we use in this paper are $a\in \{\ouralg, \text{WA*}_{\pi_{\text{baseline}}}, \text{NN}_{\pi_{\text{baseline}}}\}$, representing A* with pruning by some predicate $\pi$ (Algorithm~\ref{alg:a*} with pruning), weighted A* with the baseline predicate, and neural networks with the baseline predicate, respectively. We abbreviate $\text{WA*}_{\pi_{\text{baseline}}}$ as WA* and $\text{NN}_{\pi_{\text{baseline}}}$ as NN. 
    
    For example, for our baseline algorithm, the time taken to solve a puzzle $p$ is $t(\ouralg, \pi_{\text{baseline}}, p)$. 
    Similarly, the time it takes Algorithm~\ref{alg:a*} with pruning using a learned predicate $\pi^\circ$ is $t(\ouralg, \pi^\circ, p)$ and the number of nodes expanded is $\mathcal{E}(\ouralg, \pi^\circ, p)$. 
    
Compared to our baseline, the relative time reduction from using algorithm $a$ with learned parameter $\phi$ on the set of puzzle instances $\mathcal{P}$ is defined as the {\em time speedup}:
\begin{align}
\text{speedup}_t(a, \phi, \mathcal{P})&=\dfrac{\sum\nolimits_{p\in \mathcal{P}}t(\ouralg, \pi_{\text{baseline}}, p)}{\sum \nolimits_{p\in \mathcal{P}}{t(a, \phi, p)}}. \label{eq:timespeedup} 
\end{align}
	
\nin Similarly the {\em expansion speedup} is:
\begin{align}
\text{speedup}_{\mathcal{E}}(a, \phi, \mathcal{P})&=\dfrac{\sum\nolimits_{p\in \mathcal{P}}\mathcal{E}(\ouralg, \pi_{\text{baseline}}, p)}{\sum \nolimits_{p\in \mathcal{P}}{\mathcal{E}(a, \phi, p)}}.  \label{eq:exspeedup}
\end{align}
\nin Notice that the set $\mathcal{P}$ can consist of a single problem instance, which then defines a per-instance speedup.

Treating time speedup as an objective function, we seek an approximate solution to the optimization problem:
	\bea
	\phi^* = \argmax\limits_{\phi \in \Phi_a} \text{speedup}_{t}(a, \phi, \mathcal{P}) \label{eq:opttime} 
	\eea
	where $\Phi_a$ is a space of parameters for algorithm $a$ defined on paths and puzzle instances. For example, $\Phi_\textup{WA*}=\{1.0, 1.1, \ldots, 9.9\}$ contains possible values of WA*'s weights, while $\Phi_{\ouralg}$ contains predicates.
    Thus, given a puzzle instance set $\mathcal{P}$, we wish to find the predicate $\pi^*$ with the highest value of $\text{speedup}_t(\ouralg, \pi^*, \mathcal{P})$. 
	
	A desirable solution to the optimization problem would be found automatically and would be portable: a predicate synthesized for one puzzle should speed up search on other puzzles. 
    We also prefer the predicate to be compact and human-readable.
    Finally, we prefer the predicate to be verifiable as having no false positives so it can be used to prune the open list (Theorem~\ref{thm:fp}).

	%%%%%%%%%%%%%%%%%%%%%%%%%%%%%%%%%%%%%%%%%%%%%%%%%%%%%%%%%%%%%%%%%%%%%%%%%%%%%%%%%%%%%
	
	\section{Related Work}
	\label{sec:relatedWork} 
	Browne \cite{browne2013deductive} used deductive search to solve \textit{Slitherlink} puzzles~\cite{slitherlink} whose constraints are similar to triangle Witness puzzles. 
    They modelled each puzzle as a constraint satisfaction problem and used hand-coded rules to manually reduce the domains of each variable. 
    In our work, we aim to find rules that can speed up search automatically, rather than by hand. 
	
	Butler, Torlak, and Popovi\'{c} \cite{butler2017synthesizing} used program synthesis to find strategies for solving \textit{Nonogram} puzzles. 
    They evaluated the output of their system by comparing it to a set of documented strategies while we tested the effectiveness of our learned predicate in speeding up search.
	
	Krishnan and Martens \cite{krishnan2022synthesizing} synthesized chess tactics using ILP. 
    However, they did not test the effectiveness of their chess tactics in actual games; we test the effectiveness of our predicates on actual Witness puzzles.
	
	Kraj\~{n}ansk\`{y} et al. \cite{krajvnansky2014learning} learned predicates for pruning actions in classical planning. 
    This approach, and similar work in classical planning that pruned states from heuristic search~\cite{hoffmann2001ff, richter2010lama}, use the relaxed plan of a problem. 
    Such a relaxed plan is computed by dropping constraints from the problem.
    However, an effective constraint relaxation strategy is not obvious for Witness puzzles since removing a single constraint can substantially change the solution path. 
	
	Chen, Sturtevant, and White \cite{chen2023computing} automatically estimated the difficulty of puzzles using human-designed predicates for solving different types of Witness puzzles. 
    Our work is complementary to theirs in that automatically found predicates could improve their puzzle difficulty estimator beyond their handmade predicates. 

    Bulitko, Wang, et al. \cite{bulitko2022portability} used genetic algorithms and simulated annealing to synthesize algebraic heuristic functions that speed up search on video game maps. 
    Our work is complementary in that they synthesized heuristic functions while we hold the heuristic function fixed and attempt to synthesize predicates.
	Chen and Sturtevant~\cite{chen2021avoiding} used priority functions to avoid reopening nodes in heuristic search. 
    These priority functions were manually designed and based on $g$ and $h$, which makes it unclear how to encode a binary property of a given path. 
    We compare our approach and weighted A*, which uses priority functions of the form $g+w\cdot h$. 
	
	Botea and Bulitko \cite{botea2022tiered} used multiple expansion tiers, placeholder nodes and constraint propagation to speed up solving Romanian crossword puzzles. 
    The tiers were defined by the two types of words specific to Romanian crosswords and the number of points they contributed to the solution. 
    Their tiering mechanism is similar to our sorting of the open list with a predicate. However, our predicates are machine-learned.

	Orseau and Lelis \cite{orseauL21} extended the Levin tree search algorithm to learn a neural network for both the policy and heuristics to solve a different non-triangular type of Witness puzzle. 
    We adapt their method to our type of Witness puzzle and empirically compare its performance to our approach. More details on this approach are included in Appendix~\ref{sec:levin}.

	%%%%%%%%%%%%%%%%%%%%%%%%%%%%%%%%%%%%%%%%%%%%%%%%%%%%%%%%%%%%%%%%%%%%%%%%%%%%%%%%%%%%%
	%\subsection{Background}

	\section{Proposed Approach}
	
	We begin by presenting our predicate synthesis algorithm for learning a completability predicate, $\pi$.
    To achieve the desirable solution properties that we outlined in Section~\ref{sec:accel}, we use ILP. This enables learning human-readable predicates.  
	
	\subsection{Predicate Synthesis Algorithm}
	\label{sec:overall}
	Our overall approach is shown as the pseudocode in Algorithm~\ref{alg:unified}.\footnote{We define  $\argmax^k\nolimits_{x \in X} f(x)$ as a function that returns the $k$ elements $x_1, \ldots, x_k$ in $X$ that have the $k$ highest values of $f(x)$.} 

    Lines~\ref{al:popper} through~\ref{al:filterstep3} of Algorithm~\ref{alg:unified} comprise our machine learning algorithm for the predicates, with the puzzle sets $\mathcal{P}_{\text{train}}$, $\mathcal{P}_{\text{filter}_1}$, $\mathcal{P}_{\text{filter}_2}$, and $\mathcal{P}_{\text{filter}_3}$ used for machine learning. 
    The predicate $\pi^\circ$ is the sole output of the learning process. It is computed over each of the constraints in the puzzle to form a predicate over a partial path and the entire puzzle. 
    
	\begin{algorithm}[htbp]
		\DontPrintSemicolon
		\caption{Predicate synthesis algorithm for completability predicates. $\ouralg$ is Algorithm~\ref{alg:a*} with pruning.} 
		\label{alg:unified}
		{
			\SetKwInOut{Input}{input}
			\SetKwInOut{Output}{output\,}
			\SetKwInOut{Assert}{assert}
			\Input{training problem set $\mathcal{P}_{\text{train}}$, filter problem sets  $\mathcal{P}_{\text{filter}_1}$, $\mathcal{P}_{\text{filter}_2}$, and $\mathcal{P}_{\text{filter}_3}$, filter numbers $k_1$ and $k_2$}
			\Output{predicate $\pi^\circ$}
			%\Assert{$|\mathcal{P}_{\text{filter}_1}|<|\mathcal{P}_{\text{filter}_2}|<|\mathcal{P}_{\text{filter}_3}|$ and $|\mathcal{P}_{\text{train}}|>k_1>k_2$} 
			
			\ForEach{$p\in \mathcal{P}_{\text{train}}$}{
				$\pi^{p}\gets \textit{Popper}(p)$\; \label{al:popper}
			}  
			
			$(\pi_1, \dots, \pi_{k_1}) \gets \operatorname{argmax}^{k_1}_{\{\pi^{p}\ \mid\ p \in \mathcal{P}_\text{train}\}} \text{speedup}_{t}(\ouralg, \pi^{p}, \mathcal{P}_{\text{filter}_1})$ \; \label{al:candidateChampions}
			
			$(\pi_1, \dots, \pi_{k_2}) \gets \operatorname{argmax}^{k_2}_{\{\pi_i\ \mid\ i \in \{1, \dots, k_1\}\}}\text{speedup}_{t}(\ouralg, \pi_{i}, \mathcal{P}_{\text{filter}_2})$\; \label{al:filterstep2} 
			
			$\pi^{\circ} \gets \operatorname{argmax}_{\{\pi_i\ \mid\ i \in \{1, \dots, k_2 \}\}} \text{speedup}_t(\ouralg, \pi_{i}, \mathcal{P}_{\text{filter}_3})$\; \label{al:filterstep3}
			\Return $\pi^{\circ}$\; \label{al:return}
			
		}
	\end{algorithm}

	Each instance $p$ from the first set, $\mathcal{P}_{\text{train}}$, is used to generate positive and negative training examples for {\em Popper}.
    For each $p \in \mathcal{P}_{\text{train}}$, {\em Popper} learns a completability predicate $\pi^{p}$ (line~\ref{al:popper}).
	
	To accelerate finding high-quality predicates, we apply a filtering strategy based on the triage used by Bulitko, Hernandez, and Lelis~\cite{bulitko2021fast}. 
    To implement our version, we create a filtering set of puzzles and partition it into three disjoint subsets, $\mathcal{P}_{\text{filter}_1}$, $\mathcal{P}_{\text{filter}_2}$, and $\mathcal{P}_{\text{filter}_3}$ such that $|\mathcal{P}_{\text{filter}_1}|<|\mathcal{P}_{\text{filter}_2}|<|\mathcal{P}_{\text{filter}_3}|$ .
	
	We evaluate each of the generated completability predicates on $\mathcal{P}_{\text{filter}_1}$, choosing the $k_1$ predicates with the highest time speedup (line~\ref{al:candidateChampions}). 
	We then evaluate these $k_1$ predicates on a second, larger set of puzzle instances, $\mathcal{P}_{\text{filter}_2}$, and retain the $k_2$ predicates with the highest time speedup (line~\ref{al:filterstep2}). 	
	Finally, we evaluate the remaining $k_2$ predicates on the largest filter set of puzzles, $\mathcal{P}_{\text{filter}_3}$, and select the single highest-speedup predicate $\pi^{\circ}$ (line~\ref{al:filterstep3}). Note that $k_1>k_2$.
	
	To measure the performance of the learned predicate $\pi^\circ$, we use a novel testing set of puzzle instances, $\mathcal{P}_{\text{test}}$, which is disjoint from the training and filtering sets and thus not seen by the learning algorithm. 
    On this set we measure $\text{speedup}_t(\ouralg, \pi^\circ, \mathcal{P}_\text{test})$ and $\text{speedup}_\mathcal{E}(\ouralg, \pi^\circ, \mathcal{P}_\text{test})$. 
	
	\subsection{Learning Predicates from Puzzle Instances}\label{sec:popperdesc}
    To simplify the learning task, we first seek to learn a predicate $\pi'$ defined on a {\em single} constraint from the set of constraints contained in the puzzle $p$, plus a partial path. 
    Then, to compute $\pi^p$ in line~\ref{al:popper}, we call $\pi'$ repeatedly on each constraint from $p$. 
    If $\pi'$ returns True (incompletable) for any constraint, we set $\pi^p$ to True. 
    Otherwise, we set it to False.
	
	\begin{figure}[hbtp]
		\centering
		\includegraphics[width=0.8\columnwidth]{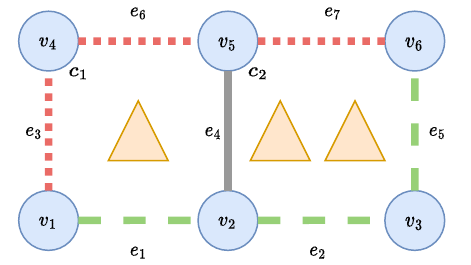}
		\caption{Figure~\ref{fig:expuzzle} with updated vertex labels.}	
		\label{fig:labeledpuzzle}
	\end{figure}
	
	We use inductive logic programming~\cite{muggleton1991inductive, cropper2022inductive} to learn each desired completability predicate.
    The ILP algorithm we use is \textit{Popper}~\cite{cropper2021learning}, which learns predicates in Prolog. {\em Popper} uses answer set programming to generate predicates based on background knowledge. It then evaluates each candidate using the positive and negative examples in the data set. If a candidate predicate satisfies a negative example, {\em Popper} prunes generalizations of the predicate from the space of candidates. If a predicate fails to satisfy some positive examples, {\em Popper} prunes specializations of the predicate from the space. This allows {\em Popper} to prune multiple candidate predicates given performance of a single candidate predicate on the training data. If {\em Popper} cannot find a predicate that satisfies all positive examples and does not satisfy any negative examples, it returns a predicate that satisfies the most positive examples while not satisfying any negative examples. We illustrate with the puzzle from Figure~\ref{fig:expuzzle}, augmented with vertex labels in Figure~\ref{fig:labeledpuzzle}.

	We supply {\em Popper} with {\em background knowledge}: predicates to use as building blocks while learning the completability predicate (Table~\ref{tab:bkbias}).
    For instance, \Verb#count(A,B,C)# specifies the number of elements in common between lists $A$ and $B$ is $C$. 
	
	\begin{table}[t]
			\caption{Predicate building blocks for the background knowledge given to {\em Popper}.}
		\centering 
		\begin{tabular}{c|c} 
			\toprule
			{\bf Predicate} & \textbf{Returns True if} \\ \midrule 
			\Verb#edge(A, B,C)# & \makecell{$A$ is an edge connecting vertices $B$ and $C$} \\ \midrule 
			\Verb#square(A,B,C)# &\makecell{$B$ is the number of triangles and\\ $C$ is the list of edges around square $A$} \\  \midrule 
			\Verb#path(A,B)# &$B$ is the list of edges in path $A$ \\ \midrule 
			\Verb#pathHead(A,B)# &$B$ is the vertex at the head of path $A$ \\ \midrule 
			\Verb#count(A,B,C)# &\makecell{the number of occurrences of list $A$\\ in list $B$ is $C$, that is $|A\cap B|=C$} \\ \midrule 
			\Verb#len(A,B)# & the length of list $A$ is $B$ \\ \midrule 
			\Verb#gte(A,B)#  &$A\ge B$ \\ \midrule 
			\Verb#greaterThan(A,B)# &$A>B$ \\ \midrule 
			\Verb#incident(A, B)# &\makecell{last vertex of path $A$ has an edge\\  incident to it in square $B$} \\ \midrule 
			\Verb#notIncident(A,B)# &\makecell{last vertex of path $A$ does not have\\ an edge incident to it in square $B$} \\ \midrule 
			\makecell{{\tt one(A), two(A),} \\ {\tt three(A)}} &$A$ is $1$, $2$ or $3$ respectively. \\ \midrule 
		\end{tabular}
		\label{tab:bkbias}
	\end{table}

	\sloppy The predicates in Table~\ref{tab:bkbias} can be used to describe a puzzle instance. 
    For example, for Figure~\ref{fig:labeledpuzzle}, \texttt{square(c1,~1, [e1,~e3,~e4,~e6])} states that the square \Verb#c1#  contains one triangle and is defined by the edges listed.
	
	To generate examples for \textit{Popper} given a puzzle instance $p$, we enumerate all paths (without vertex repetitions) between the start and the goal vertices and check to see which are solutions to the puzzle. 
    Any partial path that is not the prefix of a solution is incompletable and is thus recorded as a positive training example.
    Any other partial path is recorded as completable and a negative training example. 
    For example, the only solution to the puzzle in Figure~\ref{fig:expuzzle} is $[e_1, e_2, e_5]$. 
	Notice that the only prefixes of the solution path are $p_2=[e_1]$ and $p_4=[e_1, e_2]$, which are the two negative training examples because they are completable. 
	
	The background knowledge, a bias file controlling {\em Popper}'s operation, and the training examples for a single puzzle instance $p$ are given as inputs to {\em Popper}. {\em Popper} then attempts to learn a predicate $\pi'$ that predicts completability for a partial path and a single constraint.
	
	{\em Popper} generates predicates of increasing size, where the size of a predicate is the number of head and body predicates contained in it.  For the example in Figure~\ref{fig:labeledpuzzle}, \textit{Popper} generates and evaluates predicates until $\pi^{\text{ex}}$ is learned: 
	
	\nin{\tt $\pi^{\text{ex}}$(A,B):- square(B,D,E),path(A,C),count(E,C,F), \\ greaterThan(F,D).}
	
	\nin{\tt $\pi^{\text{ex}}$(A,B):- notIncident(A,B),path(A,D), \\ one(C),len(D,C).}
	
	The first \textit{clause} is equivalent to local constraint checking, while the second clause has false positives on some puzzle instances. Our filtering strategy in Algorithm~\ref{alg:unified} can help avoid selecting predicates such as $\pi^{\text{ex}}$, which generally yield lower speedups than predicates without false positives.

	%%%%%%%%%%%%%%%%%%%%%%%%%%%%%%%%%%%%%%%%%%%%%%%%%%%%%%%%%%%%%%%%%%%%%%%%%%%%%%%%%%%%%%
	
	\section{Evaluation} \label{sec:eval}
	
	We will now describe the hyperparameters used in our empirical study. We will then present the predicate our method learned and evaluate its speedup relative to WA* and NN. 

	\subsection{Puzzle Sets and Hyperparameters for Learning a Predicate}
	We generated each set of puzzles for training and filtering as follows. We generated $\mathcal{P}_{\text{train}}$ as a set of $500$ puzzles of sizes between $2\times 2$ to $4\times 4$. 
    We chose the number of puzzles and the puzzle sizes to generate enough training data for \textit{Popper} to learn each predicate within a few minutes. 
	
	We picked the maximum number of variables used when \textit{Popper} generates predicates to be $7$. 
    Setting the number of variables higher creates the potential to learn better predicates, but increases the learning time substantially. 
    The filter sets contained puzzles of sizes between $2\times 2$ and $5\times 5$ with $|\mathcal{P}_{\text{filter}_{1}}|=100, |\mathcal{P}_{\text{filter}_{2}}|=500$, and $|\mathcal{P}_{\text{filter}_{3}}|=2,\!500$. 
    Applying Algorithm~\ref{alg:unified}, we filtered out the top $k_1=25$ of Popper's generated predicates using $\mathcal{P}_{\text{filter}_{1}}$, then the top $k_2=5$ of those results using $\mathcal{P}_{\text{filter}_{2}}$, and then finally determined $\pi^\circ$ via $\mathcal{P}_{\text{filter}_{3}}$. 
	
	The testing set $\mathcal{P}_{\text{test}}$ contained $15,\!000$ puzzles of sizes between $2\times 2$ and $5\times 5$. 
    The test size included more puzzles to get more reliable results for $\text{speedup}_t(\ouralg, \pi^\circ, \mathcal{P}_\text{test})$ and $\text{speedup}_\mathcal{E}(\ouralg, \pi^\circ, \mathcal{P}_\text{test})$. 
   	\begin{table*}[t]
		\caption{Output of Algorithm~\ref{alg:unified}: the predicate $\pi^{\circ}$ computed over a single constraint.}
		\centering
		\begin{tabular}{c|l|l} 
			\toprule 
			& \textbf{Predicate} $\pi^{\circ}(A,B)$  & \textbf{Simplified Python} \\ 
			\midrule  
			$1$ & \makecell[l]{{\tt square(B,D,C), path(A,E),} \\  {\tt count(E,C,F),} \\ {\tt notIncident(A,B),} \\ {\tt three(D), one(F).}} & \makecell[l]{{\tt if path.head is not incident to square} \\ {\tt and puzzle.triangles[square]=3} \\ {\tt and path.edges[square]=1:} \\ { \tt   incompletable=True}} \\  \midrule 
			$2$ & Same except {\tt two(F)}.  & Same except {\tt path.edges[square]=2} \\ 
			
			\bottomrule 
		\end{tabular}
		\label{tab:predicate}
	\end{table*}
	\subsection{A Learned Predicate: $\pi^{\circ}$} \label{sec:learned}

	Learned predicates progress through triage based on how much time speedup they offer in lines~\ref{al:candidateChampions} to~\ref{al:filterstep3} of Algorithm~\ref{alg:unified}.
    $\pi_\text{baseline}$ used local constraint checking  (Section~\ref{sec:intuition}), while each $\pi$ on lines~\ref{al:candidateChampions} to~\ref{al:filterstep3} was a predicate learned automatically by {\em Popper} that was given the baseline predicate by default.
	
	To find the initial set of learned predicates, running \textit{Popper} on $500$ puzzle instances in $\mathcal{P}_{\text{train}}$ produced $489$ predicates, with $11$ runs not finding a solution in our approximate one-hour time limit. 

    On each of the successful runs, {\em Popper} took an average of about eight minutes, with a range of $[0.15, 60]$ minutes. 
    This large range reflects the difference between learning predicates for smaller puzzles of size $2\times 2$, which have fewer examples to learn from (fewer than $30$), and larger puzzles of size $4\times 4$, which have many examples to learn from (over $40,\!000$) due to there being more partial paths.
	
	Our learning algorithm for predicates, Algorithm~\ref{alg:unified}, returned the predicate shown in Table~\ref{tab:predicate} as $\pi^{\circ}$; its output is computed as a disjunction of the two clauses. To get $\pi^{\circ}$, we compute the predicate over each individual constraint in the puzzle and if it returns True for any constraints, then $\pi^{\circ}$ is True.
    The two clauses offer useful additions beyond what $\pi_\textup{baseline}$ checks. 
    They capture the idea that if a partial path leaves a three-triangle constraint square's edges without including enough of them, the path can never return there because doing so would require visiting one of the square's vertices twice.
    ~This is implemented by returning True if the path head is not incident to the square ({\tt notIncident(A,B)}, meaning the path head of $A$ is not one of $B$'s vertices), the number of triangles is $3$ ({\tt three(D)}), and the number of current intersections is $1$ or $2$ ({\tt one(F)} or {\tt two(F)}). 
	
    We manually re-implemented $\pi^\circ$ in Python to match our implementation of $\ouralg$.
    This let us call $\pi^\circ$ from Algorithm~\ref{alg:a*} without invoking Prolog, improving search speed substantially.
    
	\begin{thm}\label{thm:nofp} The learned predicate $\pi^{\circ}$ has no false positives.
	\end{thm}
Thus by Theorem~\ref{thm:fp} we can safely use this predicate for pruning and retain completeness of A* search with it.
	
	\subsection{Accelerating A* with $\pi^{\circ}$ compared to Weighted A* and Neural Networks with $\pi_\textit{baseline}$}

    \subsubsection{Weighted A*}
    A variation of A* search is weighted A*~\cite{pohl1970first}. 
In it, nodes are inserted into the priority queue by $(\pi_\textup{baseline}, g+w\cdot h, h)$, where $w$ is a weight. 

    We evaluate weighted A* with pruning using $\pi_{\text{baseline}}$ to see how its performance compares to that of Algorithm~\ref{alg:a*} with pruning using the learned predicate $\pi^{\circ}$. 
Define $\mathcal{P}_{\text{train+filter}}=\mathcal{P}_{\text{train}}\cup \mathcal{P}_{\text{filter}_1} \cup \mathcal{P}_{\text{filter}_2} \cup \mathcal{P}_{\text{filter}_3}$ as all of the problems in the training and filter sets.
We compute the weight which takes the least time to solve the puzzles in $\mathcal{P}_{\text{train+filter}}$ by sampling $100$ weights in an interval of $0.1$ between $0$ and $9.9$: 
\begin{align} w^{\circ}&=\argmax_{w\in \{0, 0.1, \dots, 9.9\}}\text{speedup}_t(\text{WA*}, w, \mathcal{P}_{\text{train+filter}}). \end{align}
The best weight found of $w^{\circ}=1.9$, whose speedup on the test set is shown in Table~\ref{tab:results}. The range of time speedups is $[0.1, 14]$ and for expansion speedups is $[0.1, 16]$. 

\subsubsection{Search with Neural  Network}
We extended the work of Orseau and Lelis \cite{orseauL21} by adding local constraint checking ($\pi_\textup{baseline}$) to their policy-guided heuristic search algorithm, PHS*. More details are described in Appendix~\ref{sec:levin}.
We trained a neural network, $\text{NN}$, for PHS* on the $500$ problems in $\mathcal{P}_{\text{train}}$ for $2.5$ hours until all the problems were solved.  We trained using the bootstrapping algorithm, which attempts to solve all the puzzles under a certain number of nodes expanded~\cite{arfaee2011learning}. 
We then solved the $15,\!000$ problems in $\mathcal{P}_{\text{test}}$ using search with the neural network ($\text{NN})$.
    The range of expansion speedups for search with the neural network was $[0.003, 511]$ and time speedups was $[3\times 10^{-5}, 8]$. 
Observe that the time speedup is about $39$ times lower than the expansion speedup. 
One explanation for this is for neural networks to perform well, they have to be trained on more puzzle instances than there are in $\mathcal{P}_{\text{train}}$ (only $500$). Indeed, preliminary results showed that a neural network trained on $50,000$ puzzles sped up search more than our predicate-accelerated search on $4$ by $4$ puzzles.

Note that the neural network has the potential to have the search take longer than the baseline.
When computing the per puzzle instance expansion speedup, the neural network had an expansion speedup of less than $1$ on $65\%$ of instances. 

      \begin{table}[htbp]
    \caption{Aggregate Results.}
        \label{tab:results}
    \centering
        \begin{tabular}{c|c|c|c}
        Speedup Metric  & $\ouralg$ & WA* & NN \\ \hline 
        $\text{speedup}_{t}$ & $6.38$ & $1.11$ & $0.02$ \\ 
        $\text{speedup}_{\mathcal{E}}$ & $6.27$ & $1.04$ & $0.77$ 
        \end{tabular}
    \end{table}
    
    The results for the speedup metrics comparing the three methods of A* with inductive logic programming ($\ouralg$), weighted A* (WA*), and neural networks (NN) are shown in Table~\ref{tab:results}. Each entry is a single ratio of totals over $\mathcal{P}_{\text{test}}$ (Equations~\eqref{eq:timespeedup} and~\eqref{eq:exspeedup}), not an average of per-instance speedups, so no standard deviations are reported.

	For $\ouralg$ accelerated by $\pi^\circ$, the range of per-instance time speedups was $[0.6, 1414]$ and of expansion speedups was $[1, 1447]$. Notice these numbers are rounded.
   
    Thus Algorithm~\ref{alg:a*} with pruning using $\pi^\circ$ never expanded more nodes than the baseline on any problem in $\mathcal{P}_{\text{test}}$. 
    When computing the per puzzle instance time speedup for each $p\in \mathcal{P}_{\text{test}}$, the predicate $\pi^{\circ}$ had a time speedup of less than $1$ (i.e., $\text{speedup}_{t}(\ouralg, \pi^{\circ}, p)<1$) on roughly a third  $(33.5\%)$ of instances. 
    However, $\pi^{\circ}$ also has the potential to significantly speed up search, and on one instance had a time speedup of $1414$ and expansion speedup of $1447$.
    We analyze this instance in Section~\ref{sec:largespeed}. 
    
\begin{figure}[t]
	\centering 
	\includegraphics[width=0.99\columnwidth]{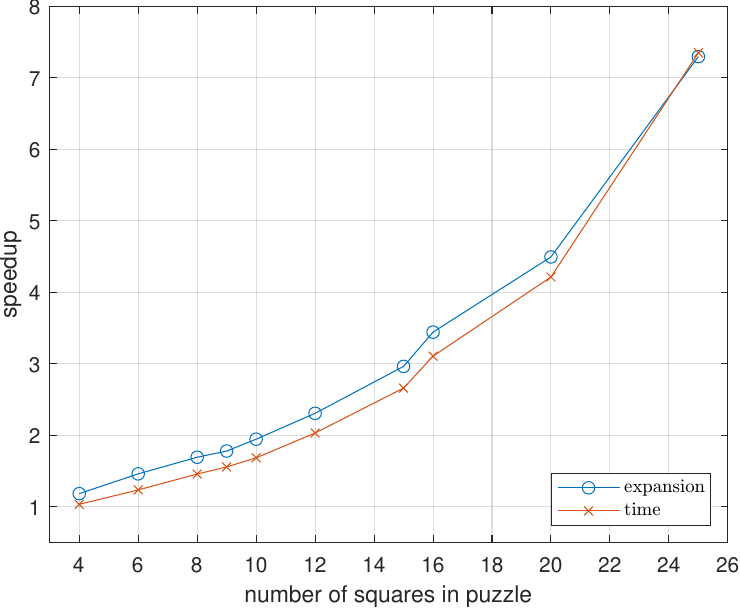} 		
	\caption{Time and expansion speedup of $\pi^{\circ}$ on $\mathcal{P}_{\text{test}}$.}
	\label{fig:exbucket} 
\end{figure}

	We also observed that the speedup afforded by using $\pi^\circ$ appears to increase with the puzzle size as seen in Figure~\ref{fig:exbucket}.

    This may occur because larger puzzles, on average, have more constraints (an average of $5.2$ on $5\times 5$ puzzles compared to $1.9$ on $2\times 2$ puzzles).
    Since predicate $\pi^{\circ}$ only applies to three-triangle constraints and since larger puzzles have, on average, more three-triangle constraints ($1.5$ on $5\times 5$ puzzles versus $0.6$ on $2\times 2$ puzzles), using $\pi^{\circ}$ might result in a higher speedup on larger puzzles. 
    
	\subsection{A Puzzle with a Large Speedup}
	\label{sec:largespeed}

\begin{figure}[htb]
	\centering
	\includegraphics[width=0.99\columnwidth]{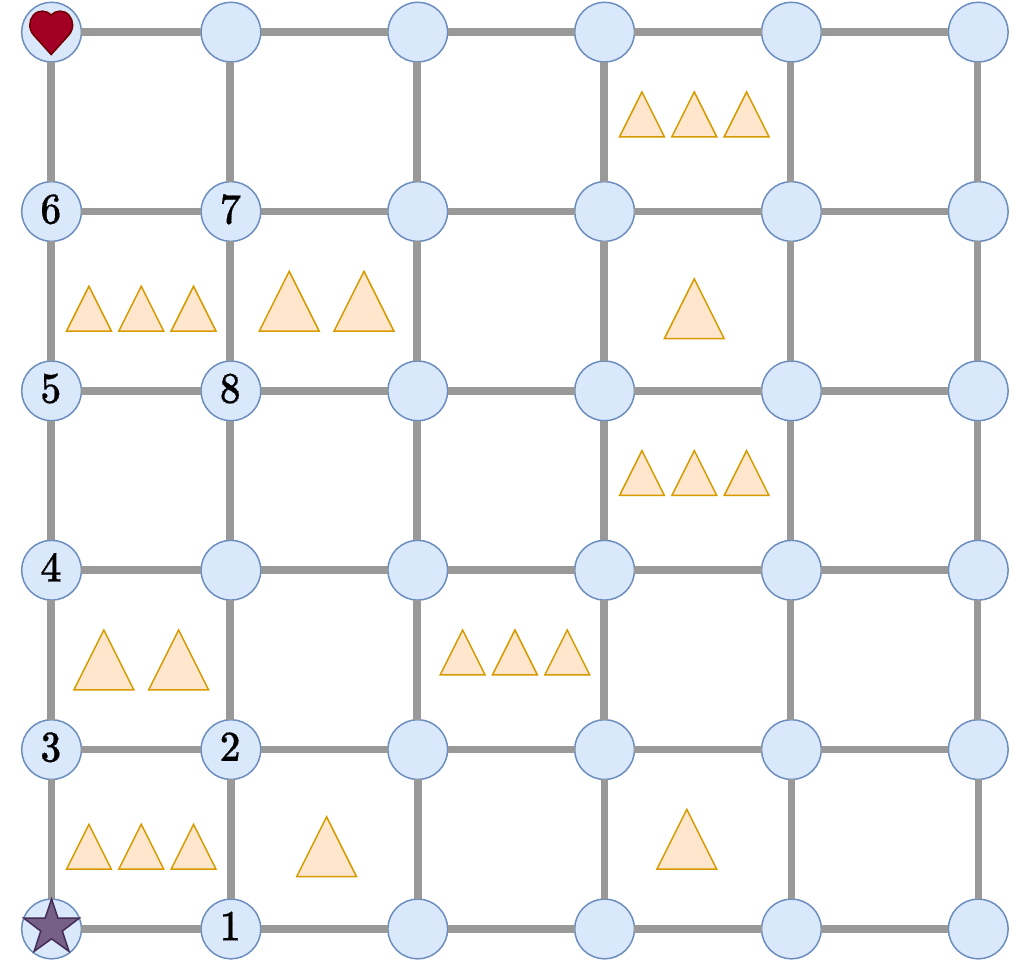}
	\caption{A $5\times 5$ puzzle. The vertices are numbered to show the partial path that pruning with $\pi^{\circ}$ forces Algorithm~\ref{alg:a*} to take. The two vertex markers $\star, \heartsuit$ mark the start and goal vertices.}	
	\label{fig:largePuzzle}
	\end{figure}

	We now show the puzzle instance from our testing set $P_{\text{test}}$ where $\pi^{\circ}$ sped up Algorithm~\ref{alg:a*} the most. 
    On it, Algorithm~\ref{alg:a*} with pruning using $\pi^{\circ}$ had a time speedup of $1414$ and an expansion speedup of $1447$. 
    The $5\times 5$ puzzle is shown in Figure~\ref{fig:largePuzzle}. 
    We show that $\pi^{\circ}$ forces Algorithm~\ref{alg:a*} with pruning to focus on partial paths whose first eight vertices visited are $[1,2,3,4,5,6,7,8]$. 
    This demonstrates how Algorithm~\ref{alg:a*} with pruning using $\pi^{\circ}$ can sometimes eliminate many partial paths from its search tree.   
	
	From the start vertex ($\star$) we can either move up to vertex $3$ or right to vertex $1$.
    Either path has one edge in common with the three-triangle square, so predicate $\pi^{\circ}$ predicts that leaving this square before satisfying the square's triangle constraint will result in an incompletable path. 
    Two possible paths to satisfy the three-triangle constraint are $[3,2,1]$ or $[1,2,3]$.
    In the first case, we are then forced to move right from vertex $1$, which would then intersect a one-triangle square twice. 
    The baseline predicate $\pi_{\text{baseline}}$ detects that this is incompletable, forcing the sequence of moves to be $[1,2,3]$.
	
	From vertex $3$ we can only move up to vertex $4$.
    From vertex $4$ we cannot move right since then the path will intersect a two-triangle square three times, which $\pi_{\text{baseline}}$ detects is incompletable.
    Thus, we must move up to vertex $5$, and our partial path so far is $[1,2,3,4,5]$. 
	
	From vertex $5$, we must move either up to vertex $6$ or right to vertex $8$. 
    In either case, we have one edge in common with the three-triangle square. 
    Predicate $\pi^{\circ}$ predicts that leaving this square without satisfying the square's triangle constraint will make the path incompletable. 
    We have two ways to satisfy the three-triangle constraint: either $[6,7,8]$ or $[8,7,6]$. 
    In the latter case, our partial path would be $[1,2,3,4,5,8,7,6]$, and our only move without repeating vertices is up to the goal. 
    However, there are still squares with triangle constraints that are not satisfied so this path is invalid. 
    Thus, no successor states of this partial path will be added to the open list. 
    Hence, the former case is the only possibility, and the starting sequence of eight moves is $[1,2,3,4,5,6,7,8]$. This is similar to an opening-move book in chess, where the first several moves are memorized ahead of time. In the same way, this sequence of moves is provided ahead of time by the predicate.
	
	\subsection{Solving Large Puzzle Instances} \label{sec:largepuzzles}
	
	We investigated how large a puzzle could be solved by Algorithm~\ref{alg:a*} with pruning using $\pi^{\circ}$ given a fixed time and memory limit. 
	We generated a set of $500$ puzzle instances of each of the following sizes: $5\times 5$, $5\times 6$, $5\times 7$, $5\times 8$, $6\times 6$, $6\times 7$, $6\times 8$, $7\times 7$, $7\times 8$, and $8\times 8$.
	
	\begin{figure}[htp]
    \centering
	\includegraphics[scale=0.5]{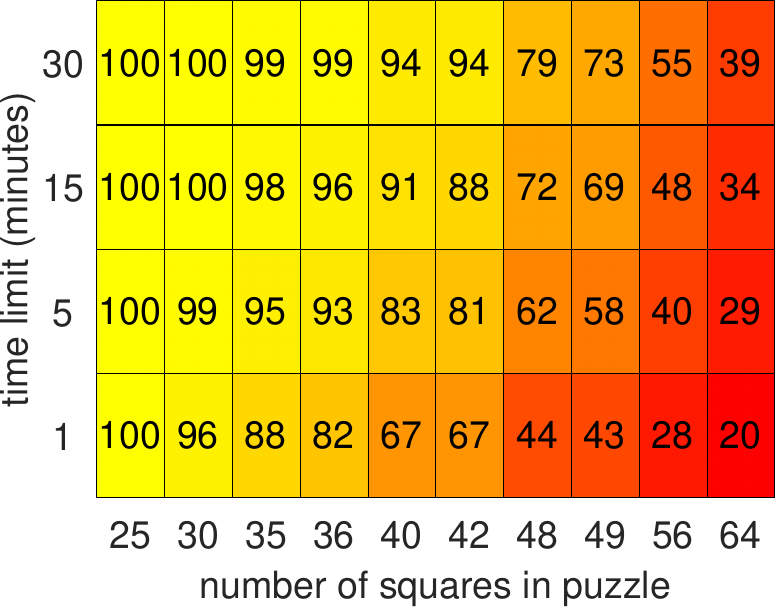} 
    
	\vspace{4mm} % having a blank line above is important for the command to work
    
	\includegraphics[scale=0.5]{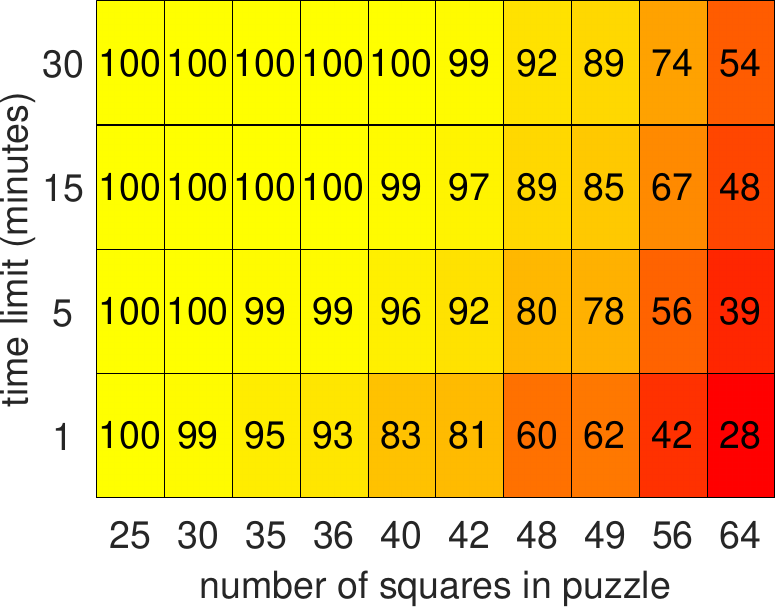}
	\caption{Rounded percentage of puzzle instances solved by Algorithm~\ref{alg:a*} with pruning using $\pi_{\text{baseline}}$ (above) and with $\pi^{\circ}$ (below).}
	\label{fig:percentageSolved} 
\end{figure}
 
We then ran Algorithm~\ref{alg:a*} with pruning using $\pi^{\circ}$ and $\pi_{\text{baseline}}$ to solve these large puzzles. 
We recorded the number of puzzle instances solved using each of the predicates under a time limit of $\{1,5,15,30\}$ minutes and memory limit of $128$ Gbytes. Figure~\ref{fig:percentageSolved} shows our results.

The large puzzle instances were difficult for the baseline search: only the $25$ and $30$ square puzzles could all be solved in under $30$ minutes. 
Algorithm~\ref{alg:a*} with pruning using $\pi^{\circ}$ solved all puzzles up to $36$ squares as well as over half of the $64$ square puzzles.
Furthermore, given a time limit of $5$ minutes, Algorithm~\ref{alg:a*} with pruning using $\pi^{\circ}$ solved approximately $84\%$ of puzzles, more than the $83\%$ that using $\pi_{\text{baseline}}$ solved given $30$ minutes. Given $1$ minute, it solved $74.4\%$ of puzzles, more than the $74.0\%$ our baseline solved in $5$ minutes. Thus, A* with $\pi^{\circ}$ solves more puzzles in less time than the baseline.

\subsection{Learned Predicate without $\pi_{\text{baseline}}$} \label{sec:humanknowledge}
We experimented with using Algorithm~\ref{alg:a*} without $\pi_{\text{baseline}}$. 
While in Section~\ref{sec:learned} we used $\pi_{\text{baseline}}$ by default in computing the speedups on the filter set, we now run Algorithm~\ref{alg:a*} with only the learned predicates. 
We re-ran Algorithm~\ref{alg:unified} by taking the $489$ predicates learned by \textit{Popper} and evaluated them on the filter sets without $\pi_{\text{baseline}}$. 
The final predicate returned by Algorithm~\ref{alg:unified}, $\pi^{'}$ contained three clauses. 
Two of the three clauses were functionally equivalent to the learned predicate $\pi^{\circ}$ in Table~\ref{tab:predicate}. 
Interestingly, the third clause, {\tt square(B,D,E), path(A,C), count(E,C,F), greaterThan(F,D)}, is \emph{equivalent} to $\pi_\textup{baseline}$ as it performs local constraint checking. This means that without any prior knowledge of completability, our approach automatically learned a predicate that captures the human-designed $\pi_{\text{baseline}}$.

Thus, the learning algorithm was flexible with respect to the human knowledge given to it. The flexibility has a simple mechanism: computing each clause of the predicate takes time, and our triage process avoids including clauses that offer insufficient speedup relative to the time they take to compute. In this particular case, the local constraint checking clause adds $26\%$ to the predicate run time on $\mathcal{P}_{\text{filter}_{2}}$ without delivering any benefits when it duplicates $\pi_{\text{baseline}}$.

\section{Open Questions \& Future Work}

\label{sec:futureWork}

We have demonstrated how an off-the-shelf ILP system with triage-based filtering (Algorithm~\ref{alg:unified}) is able to learn a predicate that substantially speeds up a baseline search algorithm in solving puzzles. 
This leads us to the following open questions. 

How much does the background knowledge provided to \textit{Popper} impact the final predicates learned? 
In particular, one can attempt to find the minimum background knowledge needed to learn the predicates.
Conversely, future work can study how to extend the grammar to learn a predicate that can speed up search even more. 

Future work can also compare our approach to other types of program synthesis such as bottom-up search or genetic algorithms.
It can also explore automatically converting predicates from Prolog to Python to gain performance benefits without the effort of a manual re-implementation.

Future work can also investigate \textit{portfolio search}, where Algorithm~\ref{alg:a*} with pruning using another learned predicate is run in parallel with Algorithm~\ref{alg:a*} and $\pi^{\circ}$.
As Algorithm~\ref{alg:a*} with pruning by $\pi^{\circ}$ is complete (Theorem~\ref{thm:nofp}), the other portfolio predicate can be used for pruning even if it has false positives. 
Thus, on some puzzle instances, an aggressively pruning member of the portfolio could yield a solution more quickly than only pruning using the more conservative $\pi^{\circ}$.

Finally, future work can investigate applicability of this approach to other puzzle types in \textit{The Witness} beyond triangle puzzles, and how it can be used to improve approaches that automatically compute the difficulty of puzzles~\cite{chen2023computing}.

%%%%%%%%%%%%%%%%%%%%%%%%%%%%%%%%%%%%%%%%%%%%%%%%%%%%%%%%%%%%%%%%%%%%%%%%%%%%%%%%%%%%%%

\section{Conclusions} \label{sec:conclusions}
Searching for solutions to puzzles can be combinatorially difficult. 
We presented an automated approach to speeding up a variation of A* via machine-learned predicates that predict the completability of a partial path. 
The human explainability of one such learned predicate can offer insights to both players and puzzle designers.
Our approach to learning such predicates can also afford the opportunity to prove that they have no false positives, which in turn allows them to be used to prune successor nodes in Algorithm~\ref{alg:a*} and preserve completeness. 
When used for pruning, one such learned predicate accelerated the baseline search by between six to seven times on our test puzzle instances over the use of baseline predicate. 
We also showed that our method can speed up search more than weighted A* and Levin tree search with neural networks~\cite{orseauL21}.

We hope that our work will encourage other game AI researchers to add these puzzles to their portfolio of game AI testbeds. To that end, we will share our code and data sets to the community upon request. 
%%%%%%%%%%%%%%%%%%%%%%%%%%%%%%%%%%%%%%%%%%%%%%%%%%%%%%%%%%%%%%%%%%%%%%%%%%%%%%%%%%%%%%

\section{Acknowledgements}  

The authors acknowledge the support of the Natural Sciences and Engineering Research Council of Canada (NSERC), including grant \#2020-06502. The authors also acknowledge support from the Digital Research Alliance of Canada, Alberta Innovates, and Alberta Advanced Education. The authors thank Andrew Cropper, Nathan Sturtevant, Jonathan Schaeffer, and Faisal Abutarab for useful discussions on the paper.

\bibliographystyle{IEEETran}
\bibliography{refs.bib}

\appendix

\subsection{Proofs}\label{sec:proofs}

	\begin{thm2} When $\pi$ has no false positives pruning partial paths that $\pi$ predicts are incompletable preserves completeness of the search.  \end{thm2}

\begin{proof} 
	Assume that pruning using $\pi$ makes Algorithm~\ref{alg:a*} return no solution for a puzzle that has a solution. Since the open list started with $v_{\text{start}}$, and $v_{\text{start}}$ is completable since the puzzle has a solution, a completable partial path must have been incorrectly pruned from $Q$. However, we assumed that $\pi$ has no false positives, which means that such incorrect pruning cannot happen. 
\end{proof} 

	\begin{thm2} The local constraint checking predicate has no false positives. \end{thm2}

\raggedbottom \begin{proof} 
	Suppose for the sake of contradiction that local constraint checking predicts a partial path $\ell$ is incompletable that is actually completable. Notice adding any edges to such a path would still result in a path that violated the triangle constraint that made local constraint checking predict True. Therefore it is impossible for $\ell$ to be completable to a solution path. 
\end{proof}

	\begin{thm2} The learned predicate $\pi^{\circ}$ has no false positives.
	\end{thm2}
\begin{proof} 
		
		Since $\pi^{\circ}$ consists of a disjunction of the two clauses in Table~\ref{tab:predicate}, it suffices to prove that each individual clause yields no false positives. In both clauses, by {\tt square(B,D,C)} and {\tt three(D)} we know that {\tt B} is a square containing a three-triangle constraint with the list of edges given by {\tt C}.
        Let the vertices incident to this square be $v_1, v_2, v_3, v_4$. By {\tt path(A,E)} we know that {\tt A} is a path with the list of edges {\tt E}. 
        Furthermore, by {\tt notIncident(A,B)}, we know that the head of path {\tt A} (a vertex in the graph) is not incident to any edge in square {\tt B}. 
        Therefore, the head of the path is not $v_1$, $v_2$, $v_3$ or $v_4$. 
		
		In the second row, given {\tt count(E, C, F)} and {\tt one(F)}, then path {\tt A} (with edges in {\tt E}) and square {\tt B} (with edges in {\tt C}) share only a single edge in common. 
        Without loss of generality, let vertices $v_1$ and $v_2$ be the two vertices along this shared edge (the edge $(v_1, v_2)$ in the red dotted line in Figure~\ref{fig:explain}).

		\begin{figure}[htbp]
			\centering
			\includegraphics[scale=1.1]{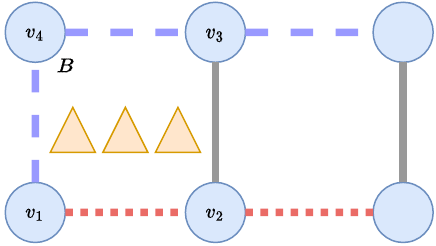}
			\caption{Portion of a puzzle with a partial path for the {\tt one(F)} case in a red dotted line and for the {\tt two(F)} case drawn in a blue dashed line. }
			\label{fig:explain}
		\end{figure}

		Assume that path {\tt A} is completable. 
        Hence square {\tt B}'s triangle constraint must be satisfied which means that path {\tt A} needs to be extended to include two more of square {\tt B}'s edges. 
        However, every pair of the three remaining edges in square {\tt B} contains an edge that either has $v_1$ or $v_2$ as one of its vertices. 
        Thus, this extension of {\tt A} would need to visit the same vertex twice, which violates rule (ii) of puzzles (Section~\ref{sec:definition}). 
        Thus, path {\tt A} is incompletable, which contradicts our assumption.
		
		The proof for the second clause is similar, however, we are given {\tt count(E, C, F)} and {\tt two(F)} instead, so we know that path {\tt A} and square {\tt B} share two edges in common in this case. 
        These two edges must include at least three vertices in square {\tt B}. 
        Without loss of generality, assume these vertices are $v_1$, $v_4$, and $v_3$ (the edges $(v_1, v_4)$ and $(v_4, v_3)$ of the blue dashed line in Figure~\ref{fig:explain}). 
		
		Assume that path {\tt A} is completable, which means that an extension of {\tt A} needs to include one more of square {\tt B}'s edges. 
        However, any additional edge will include at least one of the vertices $v_1$ or $v_3$, so this extension of {\tt A} would need to visit the same vertex twice which violates rule (ii) of puzzles from Section~\ref{sec:definition}. 
        Hence, path {\tt A} is incompletable, which contradicts our assumption.
		
		Since we have shown each clause has no false positives, their disjunction $\pi^{\circ}$ has no false positives. 
	\end{proof} 

\subsection{Search with Neural Networks} \label{sec:levin}

\begin{figure}[hbtp]
	\centering
	\includegraphics[width=0.9\columnwidth]{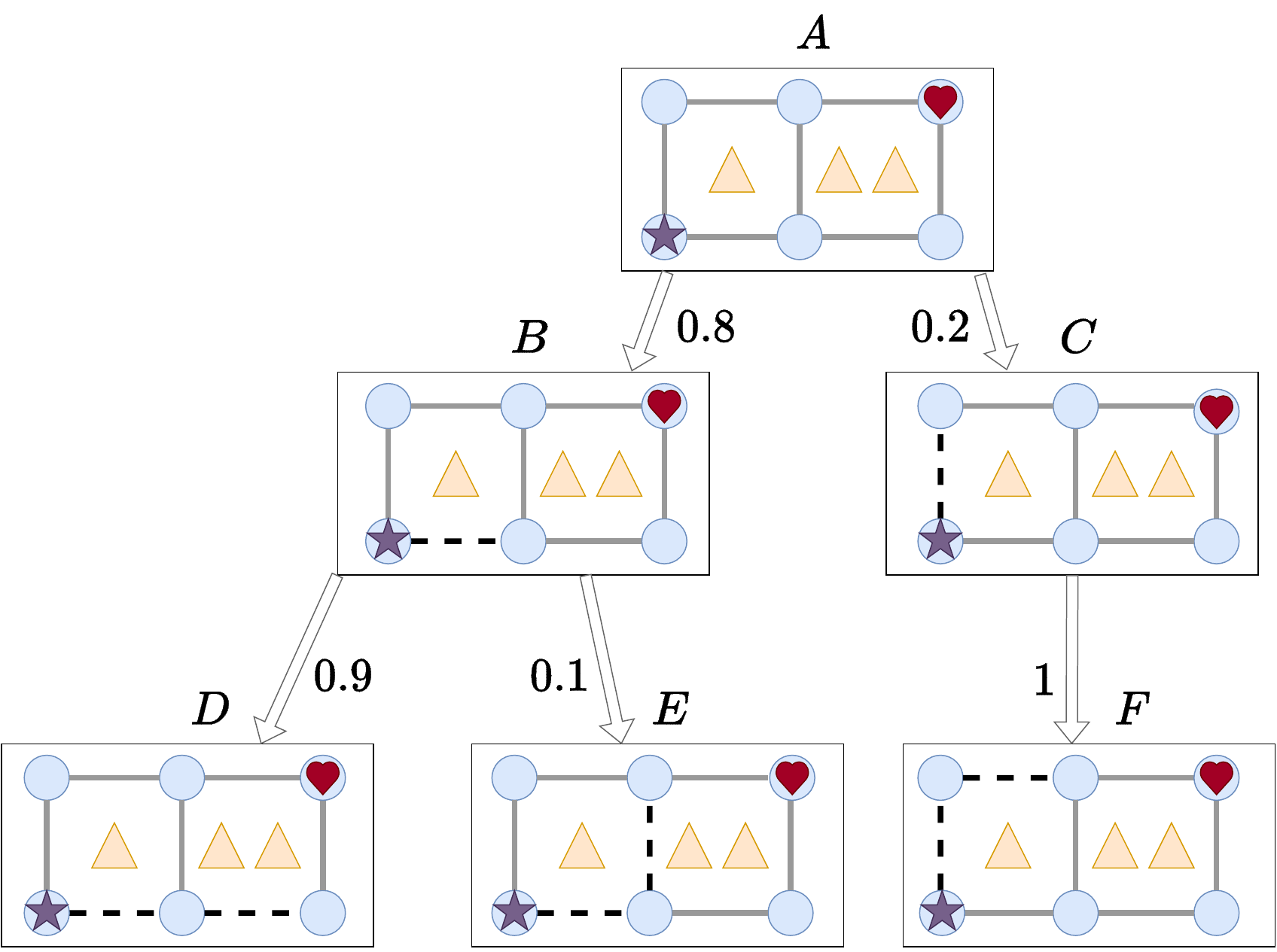}
	\caption{Search tree with a policy for solving the puzzle from Figure~\ref{fig:expuzzle}. }
	\label{fig:levin}
\end{figure}

We illustrate how search with neural networks (Levin tree search)~\cite{levin1973universal} works on an example. Consider the search tree in Figure~\ref{fig:exsearchtree} augmented with a policy that gives a conditional probability of a child node given the parent node in Figure~\ref{fig:levin}. 

Let $n_0$ be the root node of the search tree. Define the depth of a node $n$ as $d(n')=d(n)+1$ where $n'$ is a child node of $n$, with $d(n_0)=0$. For instance, in the tree in Figure~\ref{fig:levin}, $d(A)=0, d(B)=d(C)=1$, and $d(D)=d(E)=d(F)=2$. Furthermore, define the probability of getting to a given node $n$ as $p(n')=p(n|n')p(n)$ and $p(n_0)=1$ where $n'$ is a child node of $n$ ~\cite{orseauL21}. The probability of a given node is the product of all the probabilities to get from the root to that node. For example, $p(A)=1, p(B)=0.8, p(C)=0.2, p(D)=0.8\cdot 0.9=0.72, p(E)=0.8\cdot 0.1=0.08$, and $p(F)=0.2\cdot 1=0.2$. 

The ratio $d(n)/p(n)$ is called the \textit{Levin value} of node $n$. Levin tree search is a best-first search algorithm that sorts the open list by their Levin values. 

To illustrate, for the search tree in Figure~\ref{fig:levin} we begin by adding the node $A$ to the open list with its value of $\nicefrac{d(A)}{p(A)}=\nicefrac{0}{1}=0$. We then remove $A$ from the open list and add its children to the open list. The children of $A$ are $B$ and $C$ with Levin values of \begin{align*} \nicefrac{d(B)}{p(B)}&=\nicefrac{1}{0.8}=1.25\\ \nicefrac{d(C)}{p(C)}&=\nicefrac{1}{0.2}=5. \end{align*} $B$ has a lower value, therefore $B$'s children, $D$, and $E$ are added to the open list with Levin values of 

\begin{align*}
	\nicefrac{d(D)}{p(D)}&=\nicefrac{2}{(0.8\cdot 0.9)}\approx 2.8 \\ \nicefrac{d(E)}{p(E)}&=\nicefrac{2}{(0.8\cdot 0.1)}=25. \end{align*}

The open list now contains $(C, 5), (D, 2.8)$, and $(E, 25 )$. $D$ has the lowest value, therefore we expand it first. The only child of $D$ is the solution path shown in green in Figure~\ref{fig:expuzzle}, therefore, the search algorithm stops here. Notice that, similar to local constraint checking (Section~\ref{sec:intuition}), Levin tree search also does not expand nodes $E$ or $F$.

Orseau and Lelis ~\cite{orseauL21} added an additional factor to Levin tree search to learn a heuristic function from training data on solving other puzzles to guide the search. They propose policy-guided heuristic search (PHS), which adds a node $n$ to the open list with its value computed by $\frac{d(n)+h(n)}{p(n)}$. In their work they use a neural network with two outputs of both a probability function (policy) $p$ and a heuristic $h$. The actions in these heuristic search problem are moving up, down, left, and right, so the policy predicts the probability of taking these actions. The numerator considers the entire cost of the path (with the current length of the node plus the estimated cost to go), while the denominator computes the probability of a node using the same recursive formula as in Levin tree search. 

The probability of node $n$ is computed by multiplying $d(n)$ numbers together. Hence the geometric mean of these probabilities is $p(n)^{1/d(n)}$. Since the length of the entire solution path is estimated to be $d(n)+h(n)$, multiplying the geometric mean by itself $d(n)+h(n)$ times gives a probability: $$p(n)^{\frac{d(n)+h(n)}{d(n)}}=p(n)^{1+\frac{h(n)}{d(n)}}.$$ Thus in an improved version of policy-guided heuristic search since it takes into account the heuristic and estimated probability, PHS*, nodes are added to the open list with the value: $$\frac{d(n)+h(n)}{p(n)^{1+\frac{h(n)}{d(n)}}}.$$
Notice in this formula, Orseau and Lelis~\cite{orseauL21} train a neural network for the probabilities $p(n)$ and for the heuristic $h(n)$. We use their implementation augmented with local constraint checking in our experiments.

\subsection{Puzzle Sets}

As there is no existing repository of triangle puzzle instances, we generated them ourselves. We present our generators in this section with the hope that they will be used by the community to build additional triangle puzzle datasets.

\begin{algorithm}[!h]
	\DontPrintSemicolon
	\caption{Generate a puzzle instance by placing triangles randomly.}
	\label{alg:randomconstraints}
		\SetKwInOut{Input}{input}
		\SetKwInOut{Output}{output\,}
		\Input{puzzle dimensions $m,n$}
		\Output{puzzle $p$}
		\Repeat{$\ell \neq \emptyset$} { \label{alg:loop}
			$p \gets \text{empty puzzle}(m\times n)$\; \label{alg:initpuzzle}
			$\text{goal} \gets $ random location on peripheral of puzzle\; \label{alg:goal}
			$k \gets U(\{1,\dots,\lfloor \nicefrac{mn}{2} \rfloor\})$\;  \label{alg:constraints}
			$S\gets k$ randomly picked squares from the grid\;
			\label{alg:constraints2}
			\ForEach{$s \in S$}{
				$q \gets U(\{1,2,3\})$\; \label{alg:triangles}
				put $q$ triangles into square $s$\;  \label{alg:addC}
			}
			$\ell \gets$ Algorithm~\ref{alg:a*}'s solution to $p$\; \label{alg:solvable2}
		}
		\Return{$p$}\; 
		
	\end{algorithm}
	
	The first generator places triangles randomly on the grid (Algorithm~\ref{alg:randomconstraints}). In line~\ref{alg:initpuzzle} we create an empty $m\times n$ Witness puzzle.  We randomly initialize the goal location to be any vertex that is peripheral to the graph (meaning it has edge degree less than $4$) such that $v_\textup{start} \neq v_\textup{goal}$ in line~\ref{alg:goal}. The starting location, $v_\textup{start}$, is always fixed to be the bottom-left corner.
	
	In lines~\ref{alg:constraints} and~\ref{alg:constraints2} we pick the number of squares containing triangles uniformly randomly between $1$ and half of all squares (in the spirit of the challenge room maze in the actual game). We then randomly select such squares in the grid. For each of these squares we uniformly randomly assign between $1$ and $3$ triangles to it in lines~\ref{alg:triangles} and~\ref{alg:addC}. After generating the puzzle we use Algorithm~\ref{alg:a*} with $\pi_{\text{baseline}}$ to check if the puzzle is solvable (line~\ref{alg:solvable2}). If not, we restart the generation process.

	The second puzzle generator (Algorithm~\ref{alg:randomwalk}) starts with a path from the start vertex to the goal vertex and then populates the empty grid with triangles so that the path becomes a solution to the puzzle (in the spirit of~\cite{blow2011truth} work). 
	
	\begin{algorithm}[!h]
		\DontPrintSemicolon
		\caption{Generate a puzzle instance from a path.}
		\label{alg:randomwalk}
		\SetKwInOut{Input}{input}
		\SetKwInOut{Output}{output\,}
		\Input{puzzle dimensions $m,n$}
		\Output{puzzle $p$}
		$p \gets \text{empty puzzle }(m\times n)$\; \label{alg:initpuzzle2}
		$\ell \gets $ random path from start to goal \; \label{alg:rw}
		$S' \gets$ all squares in $p$ with at least one edge in $\ell$\; \label{al:edge}
		$k \gets U(\{1, \dots, |S'|\})$\; \label{al:sample} 
		$S \gets k \text{ randomly picked squares from } S'$\; \label{al:sample2} 
		\ForEach{$s\in S$}{
			$q \gets $ count $s$'s edges in $\ell$\; \label{alg:intersect}
			put $q$ triangles in square $s$\;  \label{alg:addC2}
		}
		\Return{$p$} 
	\end{algorithm}

    	To generate a puzzle set we sampled $m$ and $n$ uniformly at random between $2$ and a maximum size. Note there are fewer possible puzzles of smaller sizes (such as $2\times 2$) so the overall distribution of puzzle sizes is not uniform. All puzzles used for triage were generated with Algorithm~\ref{alg:randomconstraints}. 
    Table~\ref{tab:count} lists the distribution of tested puzzle sizes.
	
\begin{table}[!htb]
\centering
\caption{Test-set distribution and mean baseline
search costs by puzzle size.}
\label{tab:count}
\small
\begin{tabular}{@{}crrr@{}}
\toprule
Size & Instances & Expansions & Time (s) \\
\midrule
$2\times2$ & 135 & 11 & 0.0005 \\
$2\times3$ & 1,321 & 31 & 0.001 \\
$2\times4$ & 1,788 & 72 & 0.025 \\
$3\times3$ & 1,012 & 109 & 0.004 \\
$2\times5$ & 1,977 & 173 & 0.006 \\
$3\times4$ & 2,112 & 401 & 0.01 \\
$3\times5$ & 2,313 & 1,533 & 0.06 \\
$4\times4$ & 1,137 & 2,397 & 0.09 \\
$4\times5$ & 2,123 & $1.4\times10^4$ & 0.6 \\
$5\times5$ & 1,082 & $1.4\times10^5$ & 7.6 \\
\bottomrule
\end{tabular}
\end{table}

For the larger puzzles in Section~\ref{sec:largepuzzles}, we used
Algorithm~\ref{alg:randomwalk}, since it guarantees
solvability without an additional search.

\end{document}